%% file: main.tex
\documentclass{article}

 \usepackage[preprint]{neurips_2026}

\usepackage[utf8]{inputenc} 
\usepackage[T1]{fontenc}    
\usepackage{hyperref}       
\usepackage{url}            
\usepackage{booktabs}       
\usepackage{amsfonts}       
\usepackage{nicefrac}       
\usepackage{microtype}      
\usepackage{xcolor}         
\usepackage{graphicx}
 \usepackage{amsmath} 
 \usepackage{amsmath} 
  \usepackage{multirow} 
  \usepackage{bbding}
\usepackage{subcaption}
\usepackage{float}
  \usepackage[table]{xcolor}
\usepackage[most]{tcolorbox}
\usepackage{booktabs}

\title{CogBlender: Towards Continuous Cognitive Intervention in Text-to-Image Generation}

\author{%
  Shengqi Dang$^{1,2}$\quad Yi He$^{1}$\quad Jiaying Lei$^{1}$\quad Ziqing Qian$^{1}$\quad Nan Cao$^{1,2}$\thanks{corresponding author} \\
  \\$^{1}$Tongji University  $^{2}$Shanghai Innovation Institute \\
}

\def\oursName{CogBlender}
\begin{document}

\maketitle

\maketitle

\input{sec/0abstract}

\input{sec/1intro}

\input{sec/2related_works}

\input{sec/3method}

\input{sec/4exp}

\input{sec/5discussion}
\bibliographystyle{unsrt}
\bibliography{sample-bibliography}


\appendix




\newpage

\newpage

\end{document}

%% file: sec/0abstract.tex
\begin{abstract}

Beyond conveying semantic information, images also possess cognitive properties that elicit specific psychological responses from viewers, such as memory encoding or emotional reactions. Although modern text-to-image (T2I) models generate semantically coherent content effectively, they struggle to control cognitive properties (e.g., valence, memorability) and often fail to align with the user's psychological intent.
To bridge the gap, we introduce \oursName, an algorithm that enables continuous and multi-dimensional intervention on cognitive properties through a novel two-stage approach. First, we construct discrete cognition-aware rewritten prompts—variants of the input prompt that represent distinct extreme cognitive states. Second, we translate these discrete prompts into continuous control signals by interpolating within the velocity-field domain of flow-matching models. By dynamically blending the velocity fields predicted from these prompts according to the target cognitive scores, \oursName~smoothly steers the generative trajectory to realize the desired cognitive properties in the final image. Extensive experiments across four cognitive properties (i.e., valence, arousal, dominance, and memorability) demonstrate that \oursName~achieves effective cognitive intervention.

\end{abstract}

%% file: sec/1intro.tex
\section{Introduction}

When viewing an image, an observer engages in a multifaceted suite of responses, including affective appraisal, social inference, and intellectual curiosity \cite{cavanagh2011visual, gilbert2013top}. These reactions are underpinned by fundamental cognitive processes—perception, memory, and reasoning—that collectively synthesize visual stimuli into a coherent mental representation \cite{neisser2014cognitive, szczepanowski2013perception, kim2021roles}. In human-centered design, this cognitive foundation is pivotal: design objectives are increasingly framed in terms of their capacity to modulate high-level human cognition. For instance, expert designers craft images not merely for clarity, but to elicit specific cognitive reactions, such as enhancing memorability in artworks or fostering affective empathy in advertising \cite{septianto2021effectiveness, davis2023memory}.

Although recent studies have begun to investigate "cognitive image properties" or "cognitive properties of images"~\cite{GANalyze, chen2025pixelsfeelingsaligningmllms}, to the best of our knowledge, a formal and authoritative definition of this concept is still emerging. 
We define the cognitive properties of images as high-level evaluative attributes that arise from the interaction between visual stimuli and human cognitive processes, and that exhibit a measurable degree of inter-subject consistency.
Crucially, these properties are (1) multi-dimensional, inhabiting a latent psychological space of interacting facets (e.g., the multi-dimensional emotion~\cite{bradley1994measuring}); and (2) continuous, manifesting as a spectrum of intensities rather than discrete categories.

While modern text-to-image (T2I) models excel in semantic fidelity, aligning generation with high-order cognitive properties remains an open challenge. Pioneering work, GANalyze~\cite{GANalyze}, enables single-dimensional cognitive modulation by intervening in the latent space of generative adversarial networks (GANs), and more recent frameworks like EmotiCrafter~\cite{emoticrafter} achieve continuous emotion control by injecting property-specific modules into the text embedding space. Despite this progress, three critical bottlenecks persist: (1) The mapping between abstract cognitive properties and visual features is inherently non-linear and more entangled than object-level semantic alignment; (2) Cognitive traits (e.g., emotion, memorability) are studied within isolated, property-specific datasets, impeding the development of a unified model for joint multi-dimensional intervention; (3) Robust, zero-shot intervention that reliably generalizes across open-domain content remains elusive.

In this work, we present \oursName, an algorithm for \textbf{continuous and multi-dimensional cognitive property intervention} in text-to-image generation. 
Our approach enables fine-grained modulation of cognitive properties throughout the generative process while maintaining generalization across open-domain content. We instantiate \oursName~across four representative and widely-studied cognitive dimensions:
(1) {\textit{Valence~(V)}}, capturing the perceived pleasantness of an image; (2) {\textit{Arousal~(A)}}, reflecting the intensity of the elicited emotional response; (3) {\textit{Dominance~(D)}}, indicating the observer's perceived sense of agency (low: being submissive or overwhelmed vs. high: feeling influential or in control); and (4) \textit{Memorability~(M)}, a high-order cognitive attribute measuring the likelihood that an image will be retained in human memory. All four properties are psychologically meaningful, and exhibit inter-subject consistency across observers~\cite{kurdi2017introducing, kosti2017emotic, bylinskii2021memorability}. Additionally, V, A, and D are widely utilized to model emotional states~\cite{bradley1994measuring}. We model each cognitive dimension as a continuous, normalized variable within the range $[0,1]$. This formulation establishes a principled foundation for cognitive-driven image generation and provides a scalable foundation for extending \oursName~to a broader class of cognitive properties.

\oursName~takes a text prompt and a multi-dimensional cognitive score vector as input, and achieves cognitive intervention by reformulating the flow-matching process~\cite{flow-matching} within a state-of-the-art text-to-image model (e.g., FLUX.2~\cite{flux2}).
We decompose the problem into two stages: (i) constructing semantic representations for multi-dimensional extreme cognitive states, and (ii) interpolating the velocity field used in the flow-matching process to obtain continuous intervention.
Specifically, we define these extreme cognitive states as \emph{cognitive anchors}. For a given base prompt, we obtain a prompt set as the semantic representation for each cognitive anchor, by sequentially rewriting the base prompt along each cognitive dimension using a learned polarization operator, and further applying a counterbalanced rewriting strategy to reduce order bias.
To achieve continuous intervention for an arbitrary cognitive score vector, we interpolate the velocity fields predicted from different prompt sets with weights determined by the target scores.
The resulting field is integrated via the flow-matching process to generate an image that reflects the desired cognitive effects.
Extensive experiments confirm the effectiveness of our method in achieving fine-grained cognitive modulation while preserving the original semantic intent.
The main technical contributions of \oursName~are:
\begin{itemize}
    \item We define the task of {multi-dimensional continuous cognitive property intervention} in text-to-image generation. To address this, we propose \oursName, a framework capable of fine-grained cognitive modulation.
\item We decouple multi-dimensional cognitive intervention into independent single-dimensional prompt rewritings. We introduce a polarization operator for rewriting and a counterbalanced rewriting strategy to mitigate order bias, yielding a robust prompt set as the semantic representation of each cognitive anchor.   
    \item We achieve smooth and continuous cognitive intervention by interpolating among anchor-conditioned velocity fields within the flow-matching process, with weights determined by the target cognitive scores.
\end{itemize}

%% file: sec/2related_works.tex
\section{Related Works}
In this section, we review related research from two aspects: conditional image generation and cognitive properties of images.
\subsection{Conditional Image Generation}
Conditional image generation aims to synthesize images guided by diverse conditional signals, such as text prompts~\cite{flux2,sdxl}, camera poses~\cite{cameractrl}, and reference images~\cite{controlnet,IP-Adapter,T2I-Adapter}. 
Beyond these, some research has explored image generation guided by high-level attributes. 
In representative domains such as emotional image generation~\cite{emoticrafter,emoctrl,emogen,emofeedback2,uniemo,EmotionDirector,Xia2025MUSEMU,emosense}, existing works have explored both categorical and continuous control. For instance, {EmoGen}~\cite{emogen} aligns categorical emotion space with semantic space, while {EmotiCrafter}~\cite{emoticrafter} incorporates continuous Valence-Arousal values for nuanced guidance. Further advancing this, {EmoFeedback2}~\cite{emofeedback2} leverages reinforcement learning to refine emotional fidelity. Regarding image memorability, {GANalyze}~\cite{GANalyze} pioneered the use of cognitive predictors to navigate latent spaces along specific ``memorability directions.''

Existing research has primarily focused on isolated attributes, leaving the compounded effects of multi-dimensional cognitive properties unexplored. To bridge this gap, our work proposes an approach that enables continuous and multi-dimensional cognitive property intervention.

\subsection{Cognitive Properties of Images}
Cognitive properties of images refer to psychologically evaluated attributes emerging from the interaction between visual stimuli and human cognitive processes. These attributes serve as critical mediators that shape human behaviors~\cite{szczepanowski2013perception, kim2021roles}.

Emotion is one of the most fundamental cognitive appraisals elicited by images. Psychological research commonly models emotion in the valence–arousal–dominance (V-A-D) space~\cite{bradley1994measuring,kurdi2017introducing,kosti2017emotic}, representing pleasantness, intensity, and sense of control, respectively. Another property, memorability, directly determines whether visual information is retained over time and has been demonstrated to be a stable, image-level property~\cite{bylinskii2021memorability}.
Closely related to this is recallability, which refers to the ease of accurately reconstructing visual details and structural information from memory. This concept has been primarily explored in the data visualization community~\cite{VisRecall} and has yet to be fully extended to natural imagery and general visual design.
Properties such as aesthetics~\cite{MurrayMP12} and interestingness~\cite{ConstantinSIDDS21} have also been the subject of systematic study and predictive modeling, which indicates that these properties may also be algorithmically predictable. However, these dimensions are more susceptible to individual experiences, cultural nuances, and personal aesthetic priors, often leading to higher inconsistency.

Overall, we select valence, arousal, dominance, and memorability as the primary dimensions to validate our approach. However, it is designed to be dimension-agnostic and can be readily extended to additional cognitive properties.

%% file: sec/3method.tex
\section{Method}
In this section, we first introduce the problem formulation and key notation, establish the mathematical foundation underlying our approach, and then present the technical details of \oursName, describing its core components and implementation.

\subsection{Problem Formulation and Notation}

Let $p$ denote a base text prompt, and $\mathbf{s} = (s_1,\dots,s_n)\in [0,1]^n$ a target cognitive score vector, with each dimension representing a cognitive attribute (e.g., valence, arousal, dominance, memorability). The goal is to synthesize an image that remains semantically consistent with $p$ while achieving precise and continuous control over the target cognitive state $\mathbf{s}$, despite the inherently discrete nature of textual conditioning.

To parameterize cognitive control, we define the \textbf{cognitive space} as the $n$-dimensional unit hypercube $\mathcal{S} = [0,1]^n$. Ideally, each point $\mathbf{s} \in \mathcal{S}$ would correspond to a text prompt that induces the desired cognitive effect. However, due to the discrete and symbolic nature of natural language, such a continuous mapping is not directly attainable. 
To bridge this gap, we introduce a finite set of \textbf{cognitive anchors} that discretize the boundary of $\mathcal{S}$ and serve as basis states for approximation. Each anchor is defined as a binary vector
\begin{equation}
  \mathbf{a}^k \in \{0,1\}^n,\qquad k \in \{1, \dots, 2^n\},  
\end{equation}
representing an extreme cognitive configuration in which each dimension is either minimally ($0$) or maximally ($1$) expressed. We then construct semantic representations for these anchors and use them to approximate the semantics of arbitrary $\mathbf{s} \in \mathcal{S}$.
For example, in a two-dimensional valence--arousal space with base prompt $p = \text{``a valley''}$, the anchor $\mathbf{a}^1 = (0,0)$ corresponds to low valence and low arousal. A possible semantic realization is ``a desolate, misty valley rendered in cold, muted tones.''

\subsection{CogBlender}
\begin{figure*}[t]
    \centering
    \includegraphics[width=\linewidth]{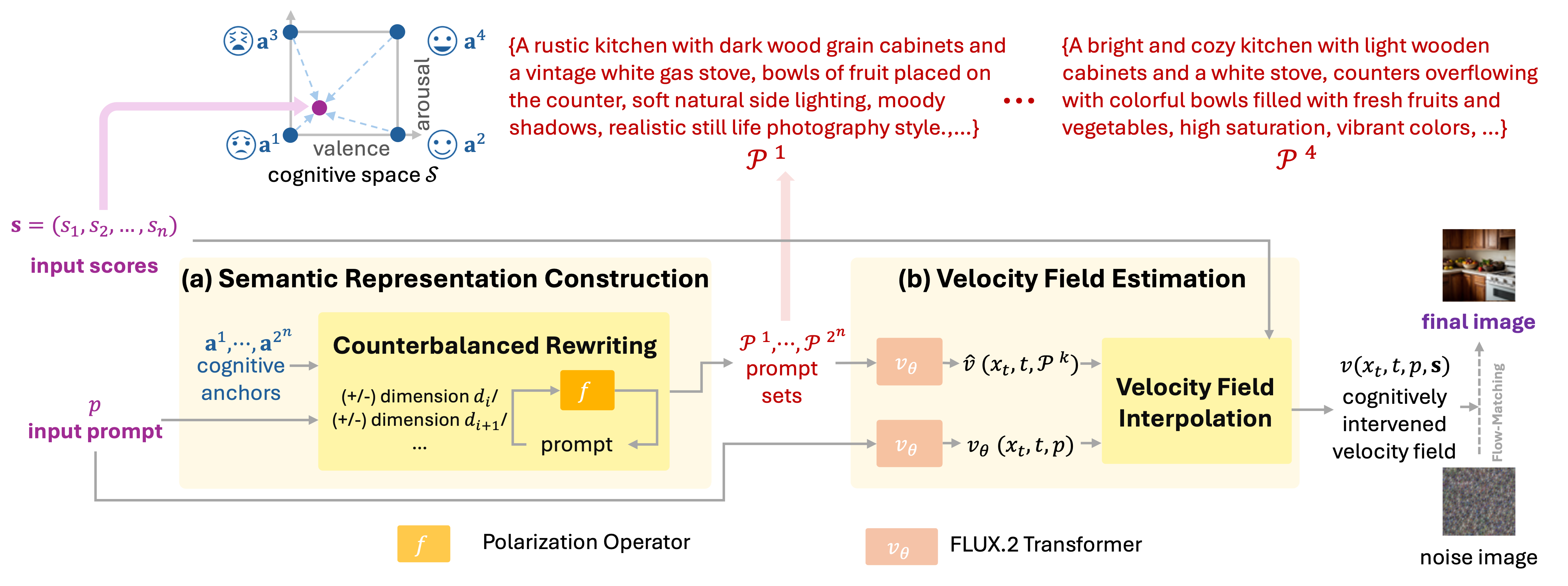}
    \vspace{-0.4cm}
\caption{\textbf{Algorithm Overview.} 
To illustrate key notations, a two-dimensional Valence-Arousal (V-A) space is depicted as a visual example.}
    \vspace{-1.2em}
    \label{fig:pipeline}
    \end{figure*}

With the above formulation, \oursName\ takes a base prompt $p$ and a target cognitive score vector $\mathbf{s}$ as input to generate images exhibiting the desired cognitive effects (Figure~\ref{fig:pipeline}). The central objective is to \textbf{adapt $p$ conditioned on $\mathbf{s}$ to enable precise cognitive intervention} while preserving the semantic fidelity and generalization ability of the underlying pre-trained model. 
However, directly encoding continuous, multi-dimensional cognitive signals into textual prompts is fundamentally challenging. First, natural language is discrete and symbolic, making it ill-suited for representing fine-grained continuous control. Second, there is a lack of supervision for jointly modeling multiple cognitive dimensions, as existing datasets are typically isolated and attribute-specific. These limitations hinder direct prompt-based approaches and motivate our alternative formulation.

To address these challenges, we propose a two-stage cognitive intervention framework that operates directly in the \emph{velocity-field space} of flow-matching models~\cite{flow-matching}, rather than relying on fragile prompt-level manipulation: (1) \textbf{Semantic Representation Construction} (Figure~\ref{fig:pipeline}(a)): For each multi-dimensional cognitive anchor $\mathbf{a}^k$, we first construct a prompt set $\mathcal{P}^k$ that serves as a robust semantic proxy of its extreme cognitive state, enabling structured decomposition of the cognitive space. (2) \textbf{Velocity Field Interpolation} (Figure~\ref{fig:pipeline}(b)): Next, instead of interpolating in the text conditioning space, we perform interpolation directly over anchor-conditioned velocity fields predicted from $\mathcal{P}^k$. This yields a unified, cognitively controlled velocity field that enables smooth, continuous, and multi-dimensional intervention during the generative process. We describe the details of each of these steps in the next.
\subsubsection{ Semantic Representation Construction}

We propose to obtain the semantic representations of cognitive anchors by rewriting the base prompt $p$, thereby avoiding the difficulty of direct multi-dimensional prompt manipulation. Our approach decomposes multi-dimensional cognitive control into a sequence of single-dimensional transformations, each applied via a learned \emph{polarization operator}. 

Specifically, the polarization operator $f_d^a$ performs controlled rewriting along cognitive dimension $d$ toward pole $a \in \{0,1\}$, corresponding to the minimal or maximal expression of that attribute. It is instantiated as an instruction-tuned Qwen3-14B model~\cite{qwen3}, trained for single-dimensional cognitive prompt rewriting. Given an input prompt $p$, the operator produces a rewritten prompt aligned with the target cognitive direction:
\begin{equation}
p' = f_d^a(p),
\label{eq:polarization}
\end{equation}
where $a=0$ ($1$) denotes the low (high) extreme of the corresponding cognitive dimension. 

However, sequential rewriting is inherently order-sensitive: different permutations of dimension-wise transformations can lead to systematically biased prompts. To mitigate this effect, we introduce a \emph{counterbalanced rewriting} mechanism that constructs a prompt set $\mathcal{P}^k$ for each cognitive anchor, thereby reducing order-induced bias and improving the robustness of the resulting semantic representation. 
Specifically, we determine the rewriting orders using a Latin square scheme. For each cognitive anchor $\mathbf{a}^k$, we cyclically vary the starting dimension and generate $n$ distinct prompts, yielding
\begin{equation}
    \mathcal{P}^k = \{p_1^k, p_2^k, \dots, p_n^k\},
\end{equation}
where each $p_j^k$ corresponds to a unique permutation that begins with dimension $j$ and cyclically traverses all remaining dimensions. Formally,
\begin{equation}
  p_j^k =
\left(
f_{d_{j-1}}^{a_{j-1}^k} \circ
f_{d_{j-2}}^{a_{j-2}^k} \circ
\cdots \circ
f_{d_1}^{a_1^k} \circ
f_{d_n}^{a_n^k} \circ
\cdots \circ
f_{d_j}^{a_j^k}
\right)(p),  
\end{equation}
where indices are taken modulo $n$ (e.g., $d_0 \equiv d_n$), and $\circ$ denotes function composition.

The resulting set $\mathcal{P}^k$ serves as a prompt-level semantic representation of the anchor $\mathbf{a}^k$ conditioned on the base prompt $p$. Although all prompts in $\mathcal{P}^k$ correspond to the same target cognitive extreme, they are obtained via different rewriting trajectories, providing diverse yet semantically consistent realizations.

\subsubsection{ Velocity Field Interpolation}

To enable continuous cognitive control from discrete prompt representations, we exploit the linear composability of velocity fields~\cite{flux2, ho2021classifier}. Instead of performing interpolation in the text conditioning space---which is known to be unstable and poorly aligned with generative dynamics—we operate directly in the \emph{velocity-field space}. This allows us to construct a unified velocity field that supports smooth and multi-dimensional cognitive intervention. Specifically, we interpolate anchor-conditioned velocity fields to obtain a cognitively controlled velocity:
\begin{equation}
v(x_t, t, p, \mathbf{s}) =
\alpha \cdot \Bigl(
\sum_{k=1}^{2^n} w_k(\mathbf{s})\, \hat{v}(x_t, t, \mathcal{P}^k)
\Bigr)
+ (1 - \alpha) \cdot v_\theta(x_t, t, p),
\label{eq:estimation}
\end{equation}
where $x_t$ denotes the noisy latent at time $t$, and $v_\theta$ is the pretrained text-to-image backbone (e.g., the diffusion transformer in FLUX.2). The term $v_\theta(x_t, t, p)$ corresponds to the base-prompt velocity, while $\hat{v}(x_t, t, \mathcal{P}^k)$ denotes the anchor-specific velocity field derived from the prompt set $\mathcal{P}^k$. The user-controllable parameter $\alpha \in (0,1]$ controls the strength of cognitive intervention relative to the base generation (we use $\alpha = 0.75$). 

The interpolation weights are defined as
\begin{equation}
w_k(\mathbf{s}) = \prod_{i=1}^{n} \big( s_i a_i^k + (1 - a_i^k)(1 - s_i) \big),
\end{equation}
which corresponds to multilinear interpolation over the vertices of the unit hypercube. Consequently, anchors that are closer to the target cognitive state $\mathbf{s}$ receive larger weights, enabling a smooth and geometrically consistent transition across the cognitive space.

To improve computational efficiency, we adopt a stochastic approximation of the anchor-specific velocity fields. At each timestep, we uniformly sample a single prompt from each set $\mathcal{P}^k$ and use its predicted velocity as an unbiased estimator:
\begin{equation}
\hat{v}(x_t, t, \mathcal{P}^k) \approx v_\theta(x_t, t, p_j^k), \qquad p_j^k \sim \mathrm{Uniform}(\mathcal{P}^k).
\end{equation}

Given the cognitively intervened velocity field, image synthesis proceeds via the standard flow-matching process. The resulting latent is then decoded into pixel space using a pretrained VAE decoder:
\begin{equation}
I = \mathrm{Decoder}\left(x_0 + \int_0^1 v(x_t, t, p, \mathbf{s})\, dt \right), 
\quad x_0 \sim \mathcal{N}(\mathbf{0}, \mathbf{I}).
\label{eq:3}
\end{equation}

For reproducibility, we provide detailed descriptions of data construction, polarization operator training, and image generation settings in Appendix.

%% file: sec/4exp.tex
\section{Experiments and Results}\label{sec:exp}

We evaluate \oursName~through two primary tasks: (1) continuous emotional image content generation (C-EICG) and (2) memorability-aware image content generation (MICG). 
Furthermore, ablation studies are performed to verify the necessity of each component. Finally, we present additional results to demonstrate the robustness and broad applicability of~\oursName.

\subsection{Experimental Setup}\label{sec:exp-setup}
\paragraph{Baselines.}
We select several baselines covering diverse paradigms of conditional generation, as summarized in Table~\ref{tab:comparison}. (1) Prompt engineering: we employ the state-of-the-art model FLUX.2~\cite{flux2} and Nano Banana 2~\cite{nanobanana2}, converting numerical cognitive scores into descriptive text (e.g., ``slightly happy, powerful'') to steer the generation process. (2) Cognition-driven generation: we compare with EmotiCrafter~\cite{emoticrafter}, a continuous emotional image generation model designed for valence-arousal control, and GANalyze~\cite{GANalyze}, an approach that manipulates the latent codes of the BigGAN~\cite{brock2018large} along learned cognition-aware directions.

\paragraph{Metrics.}
We evaluate the proposed method across three key dimensions:
(1) {Text-Image Alignment}: We employ CLIPScore~\cite{hessel2021clipscore} to measure semantic consistency between input prompts and generated images.
(2) {Visual Quality}: We adopt CLIPIQA~\cite{wang2023exploring}, a reference-free metric, to assess image quality and aesthetics.
(3) {Cognitive Fidelity}: We use V/A/D/M‑Err to quantify the accuracy of cognitive interventions. Each error is computed as the Mean Absolute Error (MAE) between the target cognitive score and the score predicted by a dimension‑specific evaluator: FindingEmo~\cite{findingemo} for valence and arousal, a Qwen3‑VL~\cite{bai2025qwen3} fine‑tuned for dominance, and MemNet~\cite{memnet} for memorability. Lower errors indicate more precise and stable control. 

\paragraph{Testset.}
To ensure a robust evaluation across diverse scenarios, we constructed a testset consisting of 100 text prompts synthesized by GPT-4o. These prompts are carefully curated to cover a wide range of semantic categories, including both human-centric scenes (e.g., portraits, social interactions) and non-human-centric subjects (e.g., natural landscapes, urban architecture, and animals).

\begin{table}[t]
\caption{Comparison of baselines and our method across key control capabilities.}
\centering
\footnotesize
\setlength{\tabcolsep}{4pt} 
\begin{tabular}{lccccc}
\toprule
\textbf{Capability} & {Nano Banana 2} & {FLUX.2} & EmotiCrafter$^{*}$ & {GANalyze$^{\dagger}$} & \textbf{CogBlender (Ours)} \\
\midrule
Text input             & \CheckmarkBold & \CheckmarkBold & \CheckmarkBold & \XSolidBrush   & \CheckmarkBold \\
Flexible dimensional control  & \CheckmarkBold & \CheckmarkBold & \XSolidBrush   & \XSolidBrush   & \CheckmarkBold \\
Continuous control     & \XSolidBrush   & \XSolidBrush   & \CheckmarkBold & \CheckmarkBold & \CheckmarkBold \\
\bottomrule
\multicolumn{6}{l}{\footnotesize \CheckmarkBold~supported, \XSolidBrush~not supported.} \\
\multicolumn{6}{l}{\footnotesize $^{*}$ Supports only two fixed dimensions (Valence and Arousal).} \\
\multicolumn{6}{l}{\footnotesize $^{\dagger}$ Supports only Valence and Memorability, and only a single dimension can be manipulated during generation.} \\
\end{tabular}
\label{tab:comparison}
\end{table}

\subsection{Experiment I: C-EICG}
We evaluate the effectiveness of \oursName~on the C-EICG task through qualitative and quantitative comparisons under both Valence-Arousal (V-A) and Valence--Arousal--Dominance (V-A-D) settings, along with a user study. For each test prompt, we synthesize images on multi-dimensional grids of cognitive scores. Specifically, we generate 2,500 images per method for the V-A task using a $5 \times 5$ uniform grid over $\{0, 0.25, 0.5, 0.75, 1\}^2$, and 2,700 images per method for the V-A-D task using a $3 \times 3 \times 3$ grid over $\{0, 0.5, 1\}^3$.

\vspace{-1.0em}
\begin{figure}
    \centering
    \includegraphics[width=1.0\linewidth]{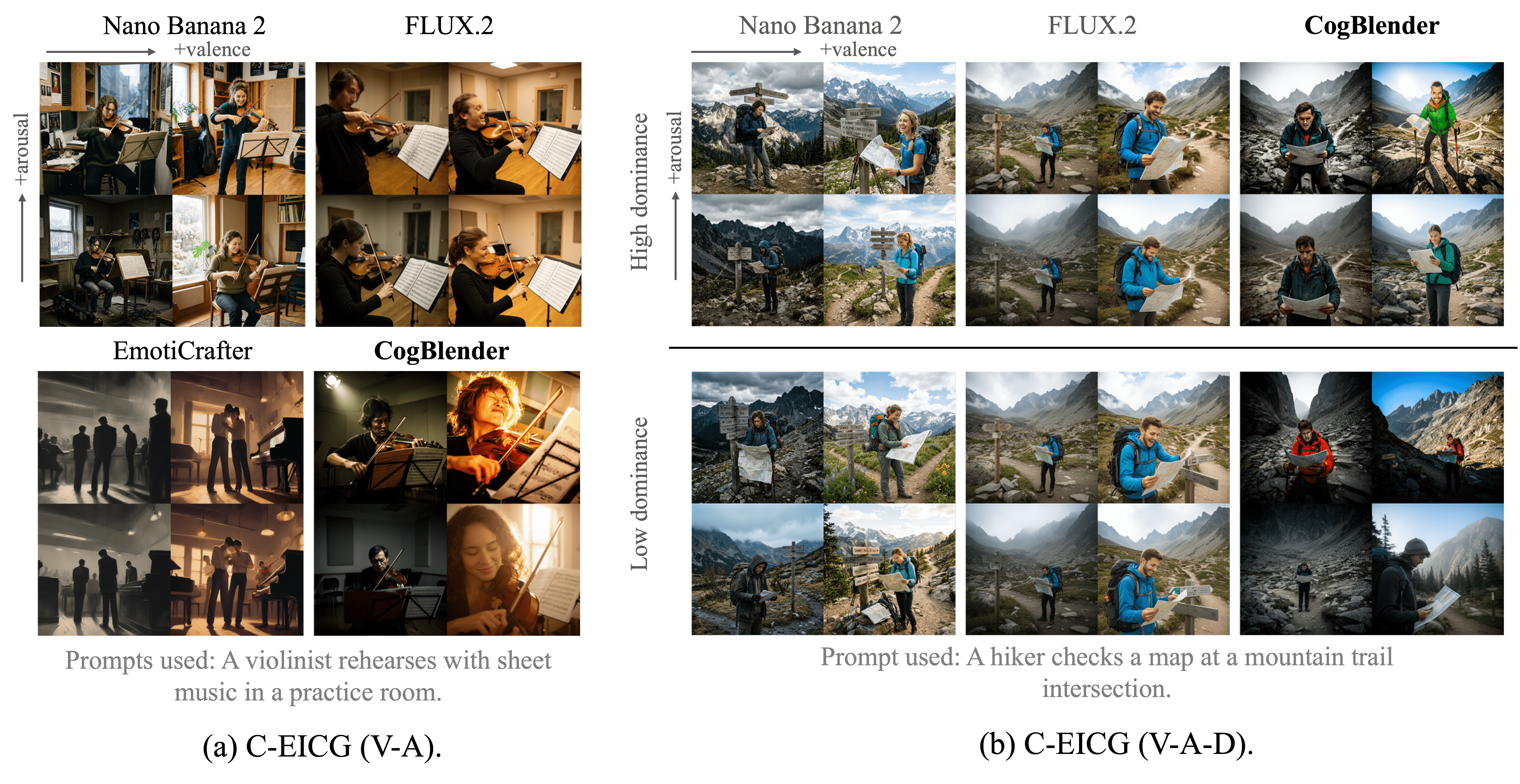}
    \vspace{-1.5em}
    \caption{Qualitative comparison on the C-EICG task.}
    \vspace{-1.5em}
    \label{fig:c-eicg}
\end{figure}
\begin{table}[t]
\vspace{-1.0em}
\centering
\caption{Quantitative comparison on C-EICG. The best and second best results are highlighted with dark blue and light blue backgrounds, respectively. 
}
\label{tab:emotion_results}
\definecolor{deepblue}{RGB}{180,180,240}      
\definecolor{lightblue}{RGB}{220,220,250}     
\resizebox{0.75\columnwidth}{!}{%
\begin{tabular}{@{}llccccc@{}}
\toprule
 & \textbf{Method} & {CLIPScore} $\uparrow$ & {CLIPIQA} $\uparrow$ & {V-Err} $\downarrow$ & {A-Err} $\downarrow$ & {D-Err} $\downarrow$ \\ \midrule

  \multirow{4}{*}{V-A}& Nano Banana 2 &  25.645  & \cellcolor{deepblue} 0.937  & 0.277 & 0.319  & -- \\
 & FLUX.2 & \cellcolor{deepblue} 26.355  & 0.885  & 0.285  & 0.323  & -- \\
 & EmotiCrafter & 17.544 & 0.861  & \cellcolor{deepblue} 0.221  & \cellcolor{lightblue} 0.304  & -- \\
 & \textbf{CogBlender (Ours)}& \cellcolor{lightblue} 25.751  & \cellcolor{lightblue} 0.917  & \cellcolor{lightblue} 0.257  & \cellcolor{deepblue} 0.299  & -- \\ \midrule
\multirow{3}{*}{V-A-D} & Nano Banana 2 & \cellcolor{lightblue} 25.774  & \cellcolor{deepblue} 0.932  & \cellcolor{lightblue} 0.312  & \cellcolor{lightblue} 0.370  & \cellcolor{lightblue} 0.357  \\
 & FLUX.2 & \cellcolor{deepblue} 26.350  & 0.886  & 0.323  & 0.372  & 0.362  \\
 & \textbf{CogBlender (Ours)} & 25.624  & \cellcolor{lightblue} 0.913  & \cellcolor{deepblue} 0.304  & \cellcolor{deepblue} 0.347  & \cellcolor{deepblue} 0.354  \\ \bottomrule
\end{tabular}%
}
\vspace{-1.0em}
\label{tab:c-eicg}
\end{table}

\paragraph{\bf Qualitative Comparison.}
As shown in Figure~\ref{fig:c-eicg}, \oursName~achieves superior emotional modulation, offering effective and disentangled control over the challenging Arousal and Dominance dimensions. Notably, even state-of-the-art closed-source models (e.g., Nano Banana 2) struggle to induce meaningful variations along these axes. Rather than relying on superficial visual adjustments such as global color shifts or lighting changes, \oursName~drives deeper semantic transformations, adaptively modifying \emph{facial expressions, body language, and spatial composition (e.g., field of view)} to align with the target cognitive state. 
In terms of text-image fidelity and image quality, EmotiCrafter frequently exhibits visual degradation and semantic drift, failing to preserve the core concepts of the prompt. In contrast, the remaining methods maintain adherence to the prompt and high visual clarity.

\paragraph{\bf Quantitative Comparison.}
As summarized in Table~\ref{tab:c-eicg}, \oursName~demonstrates robust multi-dimensional control across both V-A and V-A-D settings. Although EmotiCrafter yields a marginally lower V-Err in the V-A setting, it suffers from severe semantic collapse. In contrast, our method achieves the lowest A-Err in V-A and dominates all emotional error metrics (V/A/D-Err) in the V-A-D setting. Additionally, \oursName~maintains excellent text-image alignment (CLIPScore) comparable to FLUX.2 and Nano Banana 2.
Compared with FLUX.2, the superior visual fidelity of \oursName~can be attributed to its explicit modeling of emotional attributes, which encourages more expressive scene compositions---an effect consistent with the findings of prior work~\cite{zhu2024emotion}.

\paragraph{User Study.}
We verify the effectiveness of \oursName~through a controlled user study, measuring how well human ratings align with the target V/A/D scores and comparing against Nano Banana 2. 
We recruited 30 participants. After a tutorial on the definitions of V/A/D, participants completed the main task in three blocks, one per dimension. In each block, they viewed 30 images corresponding to 15 prompts, each generated by both \oursName~and the baseline, in randomized order, and rated only the target dimension. We normalized the ratings to [0,1] and quantified alignment between target intensities and participant ratings using mean absolute error (MAE) and Pearson’s correlation coefficient r. As shown in Table \ref{tab:user_study_c_eicg}, \oursName~achieves lower MAE and higher Pearson’s r than the baseline across all properties, indicating that \oursName~enables more reliable emotional intervention. 
\begin{table}[t]
\centering
\small
\begin{minipage}{0.9\linewidth}
\centering
\caption{User study results.}
\label{tab:user_study_c_eicg}
\resizebox{\linewidth}{!}{%
\begin{tabular}{lcccc}
\toprule
Dimension & \multicolumn{2}{c}{MAE $\downarrow$} & \multicolumn{2}{c}{Pearson's $r$ $\uparrow$} \\
\cmidrule(lr){2-3} \cmidrule(lr){4-5}
& \oursName~(Ours) & Nano Banana 2 (Baseline) & \oursName~(Ours) & Nano Banana 2 (Baseline) \\
\midrule
Dominance & 0.203 & 0.277 & 0.594 & 0.147 \\
Valence   & 0.114 & 0.123 & 0.890 & 0.884 \\
Arousal   & 0.190 & 0.237 & 0.805 & 0.549 \\
\bottomrule
\end{tabular}%
}
\end{minipage}
\vspace{-0.5em}

\end{table}

\subsection{Experiment II: MICG}
This experiment evaluates the ability of \oursName~to continuously modulate the memorability of generated images, through both qualitative and quantitative comparisons. 
For the text-to-image methods (\oursName, FLUX.2, and Nano Banana 2), we generate five images per prompt at uniformly sampled target memorability levels ranging from low to high ($\{0, 0.25, 0.5, 0.75, 1\}$, specifically), totaling 500 images per method. Due to the class-conditional nature of GANalyze, we strictly adhere to its original protocol: we randomly select 20 object categories, sample 5 random latent codes per category to establish 100 base conditions, and then manipulate each across the same five memorability intensities, also yielding 500 images in total.

\begin{figure}[t]
    \centering
    \includegraphics[width=1.0\linewidth]{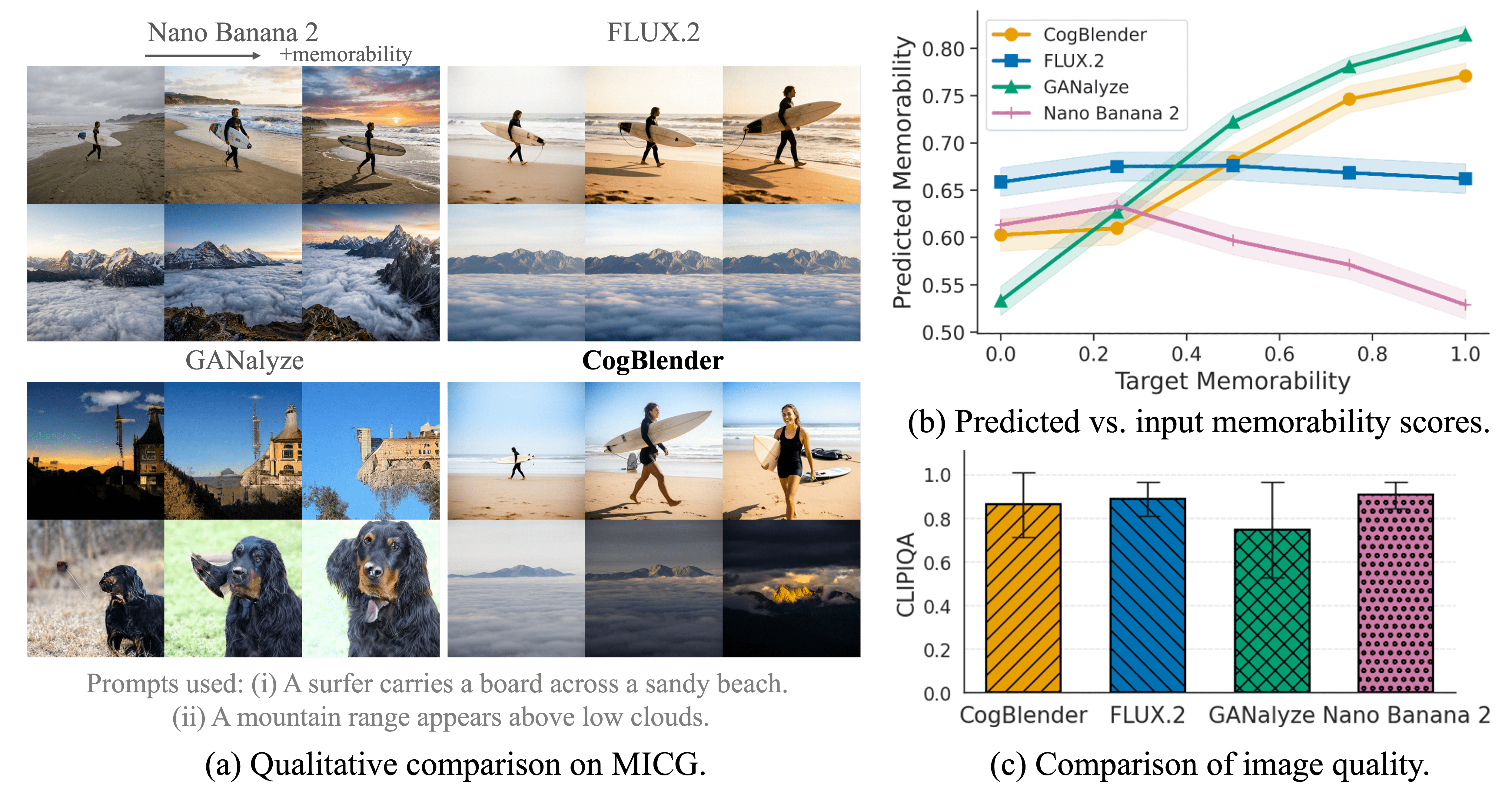}
    \vspace{-1.5em}
    \caption{Comparison results on the Memorability-Aware Image Content Generation (MICG) task.}
    \label{fig:micg}
    \vspace{-1.5em}
\end{figure}

\paragraph{Qualitative Comparison.}
Figure~\ref{fig:micg}(a) illustrates the superior semantic control achieved by \oursName~for memorability modulation. GANalyze attempts to enhance memorability primarily through low-level manipulations (e.g., zooming), which often introduce structural artifacts (e.g., the distorted dog face). FLUX.2 and Nano Banana 2 attempt to induce cognitive variation through visual feature manipulations, yet fail to produce meaningful changes.
In contrast, \oursName~preserves the core scene semantics while orchestrating high-level cognitive adjustments, including lighting, composition, and subject saliency, to naturally enhance memorability. For example, in a ``surfer on the beach'' scene, \oursName~adjusts the subject's pose (frontal vs. profile), camera perspective, and facial expression, effectively guiding the viewer's visual attention.

\paragraph{\bf Quantitative Comparison.} As illustrated in the target-predicted memorability score curves (Figure~\ref{fig:micg}(b)), \oursName~achieves a strongly positive alignment between the target and predicted memorability scores, indicating reliable and continuous cognitive intervention. In stark contrast, the prompt-engineering baselines fail to execute effective intervention.
Although GANalyze shows a steep positive slope, it suffers from severe visual quality degradation (Figure~\ref{fig:micg}(c)). Notably, all methods struggle to synthesize images with extremely
low memorability, reflecting the intrinsic difficulty of suppressing salient visual cues.

\begin{figure}[t]
\centering
\definecolor{deepblue}{RGB}{180,180,240}
\definecolor{lightblue}{RGB}{220,220,250}

\begin{minipage}[c]{0.52\linewidth}
    \centering

    \captionof{table}{Ablation on key components.}
    \label{tab:ablation_components}
    \resizebox{0.7\linewidth}{!}{%
    \begin{tabular}{@{}lcc@{}}
        \toprule
        \textbf{Method} & V-Err $\downarrow$ & A-Err $\downarrow$ \\
        \midrule
        \textbf{CogBlender (Ours)} & \cellcolor{lightblue} 0.257 & \cellcolor{lightblue} 0.299 \\
        w/o finetuned $f$ & 0.312 & 0.313 \\
        w/ fixed order (V$\rightarrow$A) & \cellcolor{deepblue} 0.237 & 0.305 \\
        w/ fixed order (A$\rightarrow$V) & 0.270 & \cellcolor{deepblue} 0.292 \\
        w/o multi-dim.\ rewriting & 0.287 & 0.309 \\
        \bottomrule
    \end{tabular}%
    }
        \vspace{2pt}
        \captionof{table}{Effectiveness of the intervention strength $\alpha$.}
    \label{tab:ablation_alpha}
    \resizebox{\linewidth}{!}{%
    \begin{tabular}{@{}lcccc@{}}
        \toprule
        \textbf{Method} & CLIPScore $\uparrow$ & CLIPIQA $\uparrow$ & V-Err $\downarrow$ & A-Err $\downarrow$ \\
        \midrule
        $\alpha = 0.25$ & \cellcolor{deepblue} 26.754 & 0.898 & 0.327 & 0.323 \\
        $\alpha = 0.5$  & \cellcolor{lightblue}26.434 & 0.912 & 0.292 & 0.312 \\
        $\alpha = 0.75$ \textbf{(Ours)} & 25.751 & \cellcolor{deepblue} 0.917 & \cellcolor{lightblue} 0.257 & \cellcolor{lightblue} 0.299 \\
        $\alpha = 1.0$  & 25.265 & \cellcolor{deepblue} 0.917 & \cellcolor{deepblue} 0.235 & \cellcolor{deepblue} 0.291 \\
        \bottomrule
    \end{tabular}%
    }
    \vspace{2pt}
    \captionof{table}{Generalization across different backbones.}
    \label{tab:ablation_backbone}
    \resizebox{\linewidth}{!}{%
    \begin{tabular}{@{}lcccc@{}}
        \toprule
        \textbf{Backbone} & CLIPScore $\uparrow$ & CLIPIQA $\uparrow$ & V-Err $\downarrow$ & A-Err $\downarrow$ \\
        \midrule
        FLUX.2 & \cellcolor{deepblue} 25.751 & \cellcolor{deepblue} 0.917 & \cellcolor{lightblue} 0.257 & \cellcolor{lightblue} 0.299 \\
        Qwen-Image & \cellcolor{lightblue} 24.740 & \cellcolor{lightblue} 0.883 & \cellcolor{deepblue} 0.234 & \cellcolor{deepblue} 0.290 \\
        \bottomrule
    \end{tabular}%
    }
\end{minipage}
\hfill
\begin{minipage}[c]{0.45\linewidth}
    \centering
    \includegraphics[width=\linewidth]{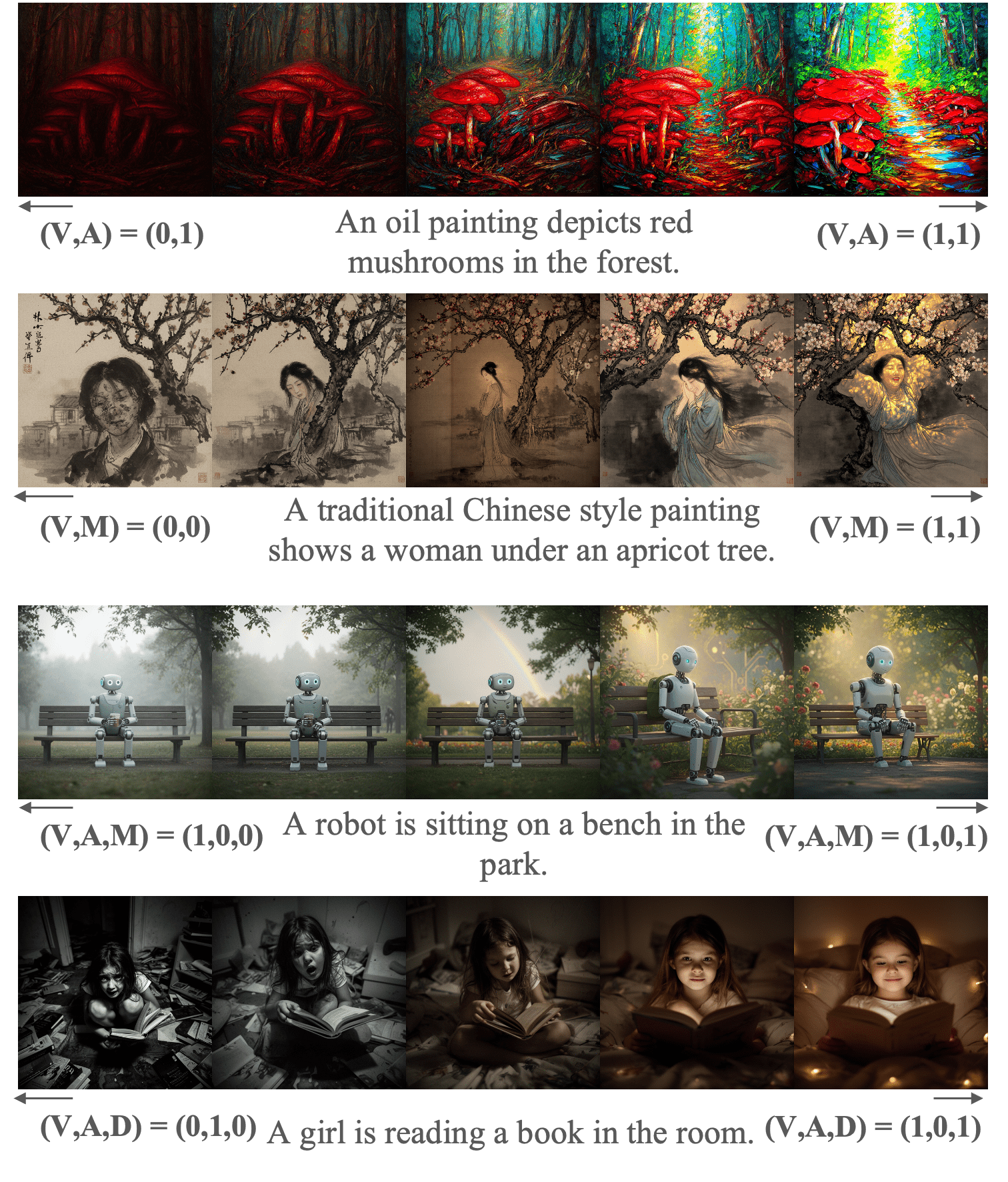} 
    \captionof{figure}{More Results.}
    \label{fig:more-res}
\end{minipage}
\end{figure}
\subsection{Ablation Study}
We conduct ablation studies on the C‑EICG task on V-A setting to assess the contribution of each key component, with results shown in Table~\ref{tab:ablation_components}, \ref{tab:ablation_alpha}, and \ref{tab:ablation_backbone}. (1)~\textbf{Finetuned polarization operator}: Replacing the finetuned polarization operator with the pretrained Qwen3 model (w/o finetuned $f$) leads to lower cognitive fidelity, indicating that cognition knowledge‑specific finetuning is essential for cognitive intervention. (2) \textbf{Counterbalanced rewriting}: Using a fixed rewriting order (either V→A or A→V)  to produce one prompt to represent the extreme cognitive state yields lower V‑Err but higher A‑Err or vice versa, confirming that the proposed counterbalanced rewriting balances all dimensions. 
(3) \textbf{Multi‑dimensional rewriting:} 
Moreover, we examine the necessity of constructing joint multi-dimensional cognitive anchors by removing these anchors and instead rewrite the prompt independently for each cognitive dimension. The resulting single-dimensional velocity fields are then linearly combined via Classifier-Free Guidance (CFG)~\cite{ho2021classifier} to generate the final image. As shown in Table~\ref{tab:ablation_components}, this approach leads to degraded cognitive fidelity, confirming that the proposed cognitive anchors are essential to properly bound the cognitive space.
(4) \textbf{Selection of $\alpha$}: As $\alpha$ increases, cognitive fidelity increases while text-image alignment drops, revealing a trade‑off. We select $\alpha=0.75$ as it strikes a good balance among these metrics. (5)~\textbf{Stochastic sampling from the prompt set $\mathcal{P}^k$}: Using stochastic sampling from the prompt set $\mathcal{P}^k$ (randomly drawing one prompt per cognitive anchor) reduces the inference time from 40.4s to 22.6s per image.

\begin{figure}
    \centering
    \includegraphics[width=\linewidth]{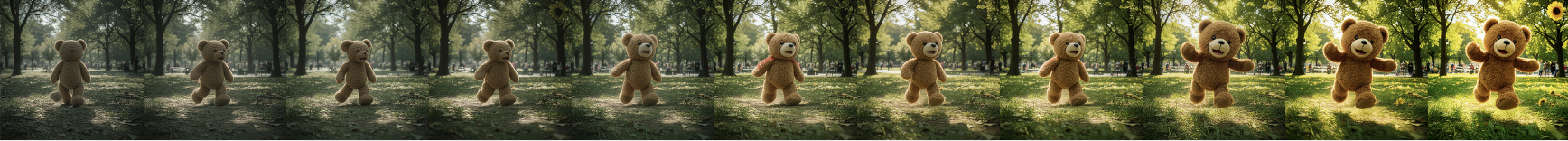} 
    \vspace{-1.0em}
    \captionof{figure}{Interpolation results along the trajectory from (V, A) = (0.3, 0.3) to (V, A) = (0.6, 0.6) for the prompt ``A teddy bear is walking in the park.''}
    \label{fig:img-inter}
    \vspace{-1.5em}
\end{figure}

\subsection{More Results}
We present additional qualitative results to demonstrate the capabilities of \oursName. 
Specifically, we show interpolation between cognitive states along specified trajectories (Figure~\ref{fig:img-inter}), generation under varying artistic styles and joint modulation of emotion and memorability (Figure~\ref{fig:more-res}). Additionally, we demonstrate that our method generalizes to another backbone (Qwen-Image~\cite{wan-image}), as shown by the results in Table~\ref{tab:ablation_backbone}.

%% file: sec/5discussion.tex
\section{Discussion and Conclusion}\label{sec:conclusion}
In this work, we introduce \oursName, an approach for generating images with specified cognitive states. The core idea is to construct cognition-aware anchor prompts via a polarization operator and to interpolate among their velocity fields during flow matching. This enables stable cognitive intervention while preserving text-image alignment. Extensive experiments demonstrate that \oursName~reliably aligns generated images with both input prompts and target cognitive scores.

While \oursName~ achieves promising results, several limitations remain. 
First, our per-step velocity interpolation incurs additional inference cost; future work could distill the interpolation mechanism directly into a feed‑forward backbone to reduce this overhead.
Second, prompt‑level rewriting provides limited control over fine‑grained visual details. Incorporating visual latent constraints could improve this, but maintaining zero-shot generalization under this setting remains an open challenge.

In conclusion, \oursName~represents a significant step toward human-centric content generation. By providing a scalable pathway to modulate how images are not just seen, but perceived and remembered, we hope to empower designers and researchers with tools that more closely align with the complexities of human cognition. We will release our code and data to facilitate further research.